\documentclass{article}
\usepackage{arxiv}

\usepackage[utf8]{inputenc}
\usepackage[T1]{fontenc}
\usepackage{graphicx}
\usepackage{latexsym}
\usepackage{epsfig}
\usepackage{times}
\usepackage{booktabs}
\usepackage{amsmath,amssymb}
\usepackage{bm}
\usepackage[caption=false]{subfig}
\usepackage{array,multirow}
\usepackage{nicefrac}
\usepackage{microtype}
\usepackage{colortbl}
\usepackage{hyperref}
\usepackage{xspace}
\usepackage{wrapfig}

\usepackage{url}
\usepackage{xcolor}

\usepackage{color,soul}
\soulregister\citet7
\soulregister\citep7

\title{Multi-Loss Weighting with Coefficient of Variations}

\author{
  Rick Groenendijk\\
  University of Amsterdam\\
  Science Park 904, 1098XH Amsterdam\\
  \texttt{r.w.groenendijk@uva.nl} \\
  \And
  Sezer Karaoglu\\
  3DUniversum\\
  Science Park 400, 1098 XH Amsterdam\\
  \texttt{s.karaoglu@3duniversum.com}\\
  \And
  Theo Gevers\\
  University of Amsterdam\\
  Science Park 904, 1098XH Amsterdam\\
  \texttt{th.gevers@uva.nl}\\
  \And
  Thomas Mensink\\
  Google Research\\
  Claude Debussylaan 34, 1082 MD Amsterdam\\
  \texttt{mensink@google.com}\\
}

\definecolor{FirstColor}{rgb}{.48,.07,.07}
\definecolor{SecondColor}{rgb}{.33,.42,.18}
\definecolor{ThirdColor}{rgb}{.07,.07,.56}
\newcommand{\fr}[1]{\textbf{\underline{\color{FirstColor}#1}}}
\newcommand{\sr}[1]{{\textbf{\color{SecondColor}{#1}}}}
\newcommand{\tr}[1]{{\textit{\color{ThirdColor}{#1}}}}
\newcommand{\logl}{\mathcal{L}} 
\newcommand{\lr}{\ell} 

\definecolor{LightCyan}{rgb}{0.88,1,1}
\definecolor{lightgrey}{rgb}{0.83, 0.83, 0.83}

\newlength{\dhatheight}

\newcommand{\ra}[1]{\renewcommand{\arraystretch}{#1}}
\newcolumntype{a}{>{\columncolor{lightgrey}}l}

\newtheorem{hyp}{Hypothesis}
\newcommand{\ourmethod}{CoV-Weighting\space}
\newcommand{\ourmethodnospace}{CoV-Weighting}
\newcommand{\ourmethodbf}{\textit{\ourmethod}}

\begin{document}

\maketitle

\begin{abstract}
Many interesting tasks in machine learning and computer vision are learned by optimising an objective function defined as a weighted linear combination of multiple losses. The final performance is sensitive to choosing the correct (relative) weights for these losses. Finding a good set of weights is often done by adopting them into the set of hyper-parameters, which are set using an extensive grid search. This is computationally expensive. 
In this paper, we propose a weighting scheme based on the coefficient of variations and set the weights based on properties observed while training the model\footnote{Source code available at: \href{https://github.com/rickgroen/cov-weighting}{https://github.com/rickgroen/cov-weighting}}.
The proposed method incorporates a measure of uncertainty to balance the losses, and as a result the loss weights evolve during training without requiring another (learning based) optimisation. 
In contrast to many loss weighting methods in literature, we focus on single-task multi-loss problems, such as monocular depth estimation and semantic segmentation, and show that multi-task approaches for loss weighting do not work on those single-tasks.
The validity of the approach is shown empirically for depth estimation and semantic segmentation on multiple datasets.
\end{abstract}

\section{Introduction} \label{sec:introduction}
In a wide variety of computer vision tasks, models are taught predictive capabilities through optimising some objective function.
While for some tasks the objective function consists of a single loss, \textit{e.g.} such as cross-entropy loss for classification, for many tasks the objective is a combination of loss functions, \textit{e.g.} a L1-loss and a Structural Similarity loss for monocular depth estimation~\cite{groenendijk2020benefit}.
In general, the final objective function $\mathcal{L}_{total}$\footnote{For clarity and consistency, the term \textbf{objective function} always denotes the final learning objective for a model / task, while the term \textbf{loss} is used to denote a single element in such a (composite) objective function.} is a linear combination of a set of loss functions $\mathcal{L}_{i}$:
\begin{equation} \label{eq:totalloss}
    \mathcal{L}_{total} = \sum_{i} \alpha_{i} \, \mathcal{L}_{i} + R(\bm\alpha),
\end{equation}
where $\bm\alpha$ denotes a set of weights and $R(\cdot)$ denotes some regularisation on these weights.

\begin{figure}
    \noindent
    \centering
    \includegraphics[width=\textwidth]{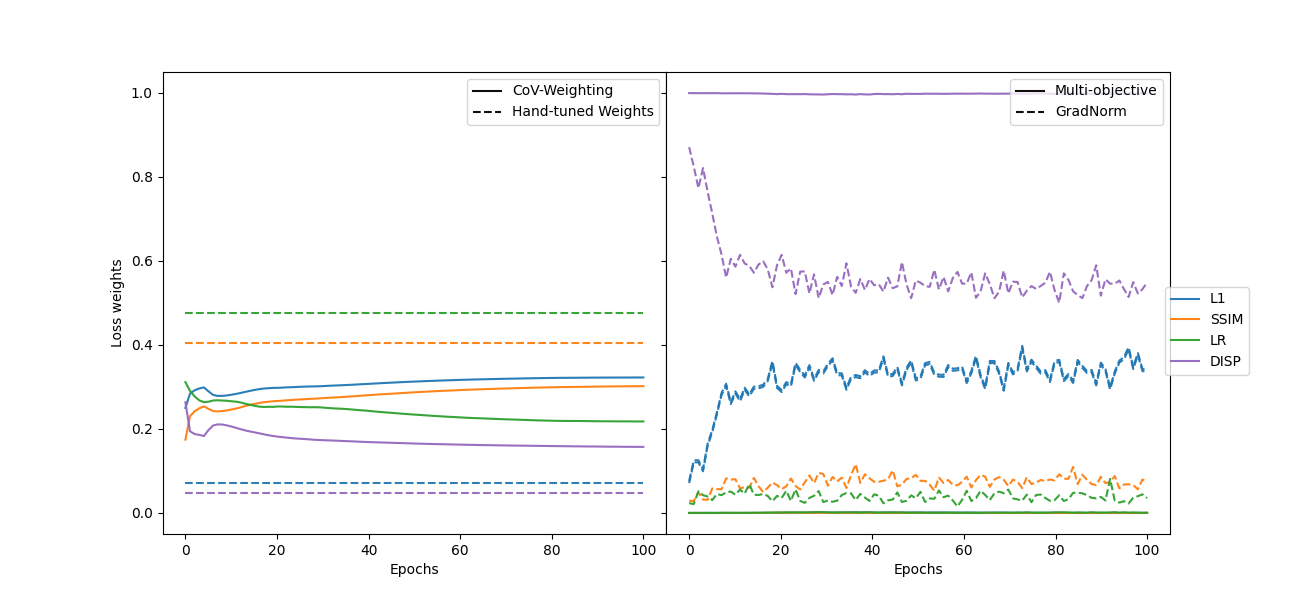}
    \caption{\textbf{(left)} Loss weights throughout training for both our method (lines) and the static optimal weights (dotted); \textbf{(right)} Loss weights throughout training for GradNorm \cite{chen2018gradnorm} (lines) and averaged loss weights of Multi-objective optimization \cite{sener2018multi} (dotted).
    }
    \label{fig:weights-cs-training}
\end{figure}

The model performance is sensitive to choosing the correct weight values $\bm\alpha$. 
In literature, the weighing of different losses is usually based on equal weighting, with the inherent assumption that each loss should contribute equally to the problem at hand, or by a hyper-parameter grid-search over the loss weights $\bm\alpha$. 
This manual tuning of loss weights is (still) prevalent, for example in segmentation~\cite{redmon2016you, he2017mask, zhang2018context}, depth estimation~\cite{godard2017unsupervised, garg2016unsupervised}, pose estimation~\cite{sridhar2015fast}, style transfer~\cite{gatys2016image, huang2017arbitrary}, and adversarial learning \cite{isola2017image}. 
This is feasible only when a small number of loss functions are combined, when more loss functions are combined. To illustrate this consider \cite{yang2018deep} who combine 40 losses, or \cite{lee2020multi} who combine 78 losses. In these cases, performing a grid-search would be too expensive computationally, and a method to set $\bm\alpha$ along with model parameters $\theta$ is desirable.

Static loss weights, \textit{i.e.} weights which are fixed through training, either by equal weighting or grid-search, might not result in optimal learning behaviour and final performance.
For example, for some learning tasks it has been shown that it is beneficial to learn easy tasks first before the more difficult tasks are introduced \cite{galama19cviu, li2017self}.
Ideally, the weights $\bm\alpha$ are adapted over time to guide the learning process.

Weighting schemes for combining multiple losses has been studied extensively in the context of multi-task learning, where multiple tasks, each with a single loss, are combined. This is appealing since conceptually task-specific information could be leveraged in related tasks to encode a shared representation \cite{caruana1997multitask, ruder2017overview}. Research in this context is tremendously broad: from architectural modifications to facilitate joint learning \cite{liu2015representation, misra2016cross, liu2019end}, to multi-task learning from domains that do not share intrinsic commonalities \cite{kaiser2017one} and from multi-task to reinforcement learning \cite{andrychowicz2017hindsight, rosenbaum2017routing}, to the use of style generation methods to learn the joint data distribution \cite{pal2018adversarial}. The main focus of this paper however is on adaptive weighting schemes for multiple losses.

In multi-task settings, the weights could be determined by, prioritising the most difficult task~\cite{guo2018dynamic}, learning the easiest tasks first~\cite{li2017self}, estimating the aleatoric uncertainty~\cite{kendall2017uncertainties, kendall2018multi}, normalising the gradients~\cite{chen2018gradnorm} or finding the Pareto optimal solution to share model capacity across the different tasks~\cite{sener2018multi}.
In contrast to these approaches this paper focus on the single-tasks, such as monocular depth estimation and semantic segmentation in which multiple losses are combined and we show that multi-task approaches do not work for these tasks.

In this paper we propose \ourmethodnospace, a single-task multi-loss weighting scheme that explicitly makes use of the statistics inherent to the losses to estimate their relative weighing. 
The method is founded on the \emph{coefficient of variation} (CoV), which is the ratio of the standard deviation $\sigma$ to the mean $\mu$ and shows the extent of variability of the observed losses in relation to the mean of the observed losses. In order to compare the CoV values of the different losses, these must have a common meaningful zero point. Therefore the CoV is estimated based on the observed loss ratio between the current observation and the mean value of that loss, instead of the direct observations of the loss. This transform different loss functions into a common scale and allow for comparisons. In ~\autoref{fig:weights-cs-training} the development of the loss weights of \ourmethod are shown throughout training. 

To validate the proposed method, extensive experimental evaluation across two tasks (depth prediction and semantic segmentation) and three datasets (KITTI, CityScapes, and PASCAL Context) is performed. 
The experiments show that~\ourmethod surpasses all multi-task loss weighting schemes and performs on par with hand-tuned weights in a single pass, with a learning rate optimised for the hand-tuned weighing.
After a hyper-parameter search over the learning rate, \ourmethod outperforms any of all current models.

\section{Method} \label{sec:method}
In this paper, we introduce a weighting scheme for single-task learning problems where the objective function is defined as a linear combination of losses. 
Each loss quantifies the cost with respect to a desired output or auxiliary objective, for example the pixel-wise L1 loss to measure the reconstruction error, a cross-entropy loss for (pixel-wise) classification, or a semantic embedding loss to capture the classes present in an image.  
The goal is to find the (optimal) set of weights $\bm{\alpha}$ for the combination of losses, \textit{c.f.} Eq. \ref{eq:totalloss}.
In order to find these weights, the following hypothesis is used:

\begin{hyp}
A loss term has been satisfied when its variance has decreased towards zero.
\end{hyp}

In other words, this hypothesis says that a loss with a constant value should not be optimised any further.
Variance alone, however, is not sufficient, given that it can be expected that a loss which has a larger (mean) magnitude, also has a higher absolute variance. Even if the loss is relatively less variant. 
Therefore, we propose to use the \textit{coefficient of variation} $c_{\logl}$ of loss $\logl$, which shows the variability of the observed loss in relation to the (observed) mean:
\begin{equation}
    c_\logl = \frac{\sigma_\logl}{\mu_\logl},
\end{equation}
where $\mu_\logl$ and $\sigma_\logl$ denote the mean and the standard deviation of loss $\logl$.
The coefficient of variation is also known as the \emph{relative standard deviation} and has the advantage that the value is independent of the scale/magnitude in which the observations are measured. This is relevant, given that we known that different losses act on different scales.
Coefficient of variation decouples  the  loss  magnitude  from  loss weighting, so a loss with small magnitude may still be relatively impactful when it is complex and variant; a bigger loss that hardly values across training examples is assigned less weight.

\subsection{Loss Ratios for Coefficient of Variation}
A requirement for computing a correct coefficient of variation is that the observations are derived from a ratio-scale, that is with a unique and non-arbitrary zero value. Then, it allows to fairly compare the uncertainty between different series of measurements, even with different magnitudes. 

Not all loss functions are measured on such a ratio-scale. 
Therefore, we propose to use the \emph{loss-ratio} $\lr$ as measurement, instead of the loss value itself, which we define as:
\begin{equation}
    \lr_t = \frac{\logl_t}{\mu_{\logl_{t\!-\!1}}},
\end{equation}{}
where $\logl_t$ is the observed loss value at time step $t$, and $\mu_{\logl_{t\!-\!1}}$ is the mean of the observed losses up to time $t-1$. The loss-ratio has also been used in multi-task loss weighting setting. 
In \cite{chen2018gradnorm} the ratio of the current observation with the first loss measurement is used ($\tfrac{\logl_t}{\logl_0}$), which we found to yield noisy and initialisation dependent ratios.
In \cite{liu2019end} the ratio of the current observation and the previous observation is used ($\tfrac{\logl_t}{\logl_{t\!-\!1}}$), which we found to yield unstable ratios. Our ratio is more robust and experimentally we explore the effect of using a decay mean instead of the mean over all observations.

The loss ratio $\lr$ has the same meaningful zero point across different loss functions: when $\logl_t$ is zero $\lr_t = 0$. Moreover it is a relative comparison of two measurements of the loss statistic. The weight $\alpha_{it}$ is based on the coefficient of variation of the loss-ratio $c_{\lr_i}$ for loss $\logl_i$ at time step $t$:
\begin{equation}
    \alpha_{it} = \tfrac{1}{z_t} \ c_{\lr_{it}} = \tfrac{1}{z_t} \ \frac{\sigma_{\lr_{it}}}{\mu_{\lr_{it}}},
\end{equation}
where $z_t$ is a normalising constant independent of $i$: $z_t = \sum_i c_{\lr_{it}}$. This ensures that $\sum_i \alpha_{it}=1$, which is important to decouple the loss weighting from the learning rate.

There are two forces that simultaneously determine the value of the loss weights:
\begin{itemize}
    \item The loss weights increase when the loss ratio $\lr_{it}$ decreases, that is when the loss $\logl_{it}$ is below the mean loss $\mu_{\logl_i}$. This encourages losses that are learning quickly, and dampens the effect on high outliers on the magnitude of the loss.
    \item The loss weights increase when the  standard deviation over the history of loss ratios $\sigma_{\lr_{it}}$ increases. This ensures that more learning occurs when a loss-ratio is more variant. That is, when a particular objective has historically been more challenging, this term makes the cost function more powerful.
\end{itemize}{}

\subsection{Robust Estimation} \label{subsec:robust-estimation}
Using \ourmethod the loss weightings are inferred directly from the history of the observed loss values. 
To estimate the loss-ratio and the coefficient of variation robustly, we use Welford's algorithm \cite{welford1962computingvariance}, an online estimate, to track the mean of $\logl$, and the mean and standard deviation of $\lr$ using the following update rules:
\begin{align}
    \mu_{\logl_t} &= \left(1 - \tfrac{1}{t} \right)\mu_{\logl_{t\!-\!1}} + \tfrac{1}{t} \logl_t, \label{eq:update-mean}\\
    \mu_{\lr_t} &= \left(1 - \tfrac{1}{t} \right)\mu_{\lr_{t\!-\!1}} + \tfrac{1}{t} \lr_t, \quad \textrm{and}\\
    \bm M_{\lr_t} &= \left(1 - \tfrac{1}{t} \right) \bm M_{\lr_{t\!-\!1}} + \tfrac{1}{t} \left( \lr_t - \mu_{\lr_{t\!-\!1}} \right)\left( \lr_t - \mu_{\lr_{t}} \right),
\end{align}{}
the standard deviation is then given by $\sigma_{\lr_t} = \sqrt{\bm M_{\lr_t}}$.
Assuming converging losses and ample training iterations, the online mean and standard deviation converge to the true mean and standard deviation of the observed losses over the data.

A potential downside of this approach is that variance over time of a loss is smoothed out. Therefore we also experiment with decaying online estimates, then $t$ is a fixed factor, \textit{e.g.} 20 or 100, to weight the aggregated previous observations and the current observation.

\section{Relation to Other Methods}
In this section \ourmethod is compared in-depth to three multi-task loss weighting methods: Gradient normalisation (GradNorm) \cite{chen2018gradnorm}, Multi-objective optimisation \cite{sener2018multi}, and Uncertainty Weighting \cite{kendall2018multi}. \autoref{tab:methods-overview} summarises the main differences between the multi-task loss weighting methods and \ourmethod.
\begin{table}[t!]
    \centering
    \setlength{\tabcolsep}{1.5pt}
    \resizebox{0.6\textwidth}{!}{
        \begin{tabular}{@{\extracolsep{10pt}}lcc@{\extracolsep{4pt}}}
            \toprule
            \textbf{Method} & \textbf{Definition} $\alpha_{i}$  & \textbf{Main Property} \\ \addlinespace[2pt]
            \hline \addlinespace[2pt]
            Uncertainty Weighting \cite{kendall2018multi}     & $\tfrac{1}{\sigma_i^2} + \tfrac{\log \sigma_{i}}{\mathcal{L}_{i}}$ & jointly learned \\ \addlinespace[2pt]
            Multi-objective \cite{sener2018multi}   & $\sum_i \alpha_i \nabla_{\theta_{\textrm{s}}} \mathcal{L}_i = 0$ & gradient based \\ [2pt]
            GradNorm \cite{chen2018gradnorm}        & $\tfrac{{\mathcal{L}_i(t)}/{\mathcal{L}_i(0)}}{g_i(t)}$ & separate loss \\ \addlinespace[2pt]
            \hline \addlinespace[2pt]
            \ourmethod                              & $\tfrac{\sigma_{\lr_{i}}}{\mu_{\lr_{i}}}$ & observed \\
            \bottomrule
        \end{tabular}
    }
    \caption{Overview of the different weighting schemes considered in this paper, with property and definition of $\alpha_i$. See text for details.
    }
    \label{tab:methods-overview}
\end{table}

\paragraph{Uncertainty Weighting~\cite{kendall2017uncertainties, kendall2018multi}.}
Uncertainty Weighting models the homoscedastic aleatoric uncertainty in multi-task settings. While homoscedastic noise is not dependent on the input data, it might be task-dependent. The observed task-loss $\mathcal{L}_i$ is seen as an observation from a Gaussian distribution $\mathcal{N}(\mathcal{L}_i; \sigma_i)$. The final objective minimises the log-likelihood of these Gaussian distributions:
\begin{equation}
    \mathcal{L}_{total} = \sum_i \frac{\mathcal{L}_{i}}{2 \sigma_{i}^{2}} + \log \sigma_{i} ,
\end{equation}
where the variance $\sigma^{2}_{i}$ is learned jointly with the model parameters. It can be shown, however, that when the optimal variance could be used, Uncertainty Weighting results in a parameter less log-loss: $\sum_i \log(\mathcal{L}_i)$. This means the smallest loss has the most impact on the gradient.

Uncertainty Weighting uses the homoscedastic (data independent) noise, in the single-task multi-loss setting this might be unsuitable since the observational noise over the same output cannot be used to weight different losses.
Moreover, in Uncertainty Weighting the loss weights increase when the observational noise in the outputs decreases. This naturally happens when the model is presented with more training data. In contrast \ourmethod uses the variance in the observed losses throughout training and when the noise in a loss ratio increases the loss weight also increases.

As opposed to other methods, Uncertainty Weighting, in the original formulation, does not satisfy $\sum_{i} \alpha_{i} = 1$, and hence there is a coupling between the loss weights (which can be arbitrarily) and the global learning rate. In \cite{kendall2018multi} the global learning rate is annealed by a power law to counteract the exponential increase of loss weights. Unfortunately this complicates direct comparison against other loss weighting methods because it is exceedingly difficult to fairly control for global learning rate.

\paragraph{GradNorm~\cite{chen2018gradnorm}.}
In GradNorm \cite{chen2018gradnorm} gradient normalisation is suggested to balance the gradient norms for each task at each training iteration step at a chosen layer $\mathcal{W}$. This layer is often the last shared layer between the different tasks. The authors argue that gradients are ideally balanced at this shared layer. In the single-task multi-loss setting, all layers are shared and hence GradNorm is computed at the output layer.

To balance multiple loss terms, the weights $\bm{\alpha}$ are learned along the model parameters, with a separate loss function. It can be derived, however, that the optimal weight values $\bm{\alpha}$ (before normalisation) equal to:
\begin{equation}
    \alpha_i \propto \frac{{\mathcal{L}_i(t)}/{\mathcal{L}_i(0)}}{g_i(t)},
\end{equation}
where $g_i(t)$ denotes the norm of the gradient of the parameters with respect to the loss $\mathcal{L}_i$ at the shared layer, hence the name GradNorm.
In other words, the loss ratio, between the loss at time t and the first loss at time 0, is divided by the current norm of the gradient.
This is counter intuitive for single-task multi-loss learning, given that their loss ratio is an \emph{inverse} measure: the smaller the value, the better the loss is training. So better performing losses are slowed down during training compared to difficult losses (where $\mathcal{L}(t) \approx \mathcal{L}(0))$, when they have an equal gradient norm.

Compared to the proposed loss ratio, in GradNorm the loss ratio is based on the initial loss value. When the network has just been initialised, this can be a poor estimate to measure the velocity with which the network learns for a specific loss. This is solved in \ourmethod by using the loss ratio between the current loss and the mean of the observed losses. 

\paragraph{Multi-Objective Optimisation~\cite{sener2018multi}.}
The authors of \cite{sener2018multi} state that while multi-task learning can, in general, be beneficial for all tasks, some individual tasks could compete for model capacity. That is, the shared encodings may not be equally informative for all tasks. Hence, the authors are interested in finding Pareto optimal solutions for tasks using the Frank Wolve algorithm \cite{jaggi2013revisiting}. A potential issue in single-task learning for multi-objective optimisation is that single-task learning is inherently a single-objective optimisation. That is, the optimal solution cannot be assumed to be a Pareto optimum between the different loss functions. It is expected that dealing with auxiliary losses will be especially challenging for this method.

\section{Experiments} \label{sec:experiments}
In this section, the proposed method is evaluated on two distinct scene understanding tasks: Depth estimation on KITTI \cite{Geiger2012CVPR, Geiger2013IJRR} and CityScapes \cite{Cordts2016Cityscapes}, and semantic segmentation on the PASCAL Context dataset~\cite{mottaghi2014role}.
The purpose of these experiments is two-fold: 
First, to compare the proposed dynamic weights to a set of static weights (equal weighting or hand tuned). 
Second, to test the proposed method against three dynamic multi-task loss weighing approaches.

To fairly compare the different methods, our code includes the proposed weighing scheme and implementations of the baseline methods. 
The dynamic baselines that are used are Uncertainty weighing \cite{kendall2018multi}, GradNorm \cite{chen2018gradnorm}, and Multi-objective optimisation \cite{sener2018multi}. 
GradNorm requires an additional hyper-parameter as a form of temperature scaling on the loss weights (before normalisation).
Preliminary experiments show that the performance is sensitive to this value; using grid-search it is set to 1.5 for all experiments. It should be noted that careful tuning of this parameter is required before performance is satisfactory, unlike with the other baselines that are used.

\begin{table}[t]
    \centering
    \setlength{\tabcolsep}{2\tabcolsep}
    \begin{tabular}{@{\extracolsep{4pt}}lccc@{\extracolsep{4pt}}}
    \toprule
    & ARD & RMSE log & $\delta < 1.25$ \\ 
    \cline{2-3} \cline{4-4}
    & \multicolumn{2}{c}{\small{lower}} & \multicolumn{1}{c}{\small{higher}} \\ 
    \midrule
    \textbf{full history}           & \underline{0.1988} & 0.3118 & 0.6988 \\
    \midrule
    \textit{t} = 20                 & 0.2159 & 0.3479 & 0.7063 \\
    \textit{t} = 100                & 0.2095 & 0.3498 & 0.7050 \\
    \textit{t} = 1000               & 0.2037 & \underline{0.3009} & \underline{0.7084} \\
    \bottomrule
    \end{tabular}
    \caption{The effect of setting \textit{t} to a factor that controls how much the updates of the statistics are effected by previous statistics. A \textbf{full history} is shown in e.g.~\autoref{eq:update-mean}, in which \textit{t} is the iteration count. Alternatively, \textit{t} could be set to a static number to allow for a fixed decay. There is no clear benefit of using decay, hence the full history of statistics is used in further experiments.}
    \label{tab:preliminary-cityscapes-statistics}
\end{table}

\subsection{Depth Estimation}
In these experiments, we learn to estimate depth from left/right image pairs by means of photo-metric reconstruction, using an estimated disparity map; depth can be inferred by warping the disparity map using the camera intrinsics.
We follow the network architecture and objective functions of \cite{godard2017unsupervised}.
The objective function combines the L1 loss, Structural Similarity loss (SSIM), left-right consistency loss (LR), and disparity gradient loss (DISP) to train a single network.
Following \cite{godard2017unsupervised}, the hand-tuned weights are set to $\{\alpha_{L1}=0.15, \alpha_{SSIM}=0.85, \alpha_{LR}=1.0,$ $\alpha_{DISP}=0.1\}$, which are normalised such that $\sum_{i} \alpha_{i}=1.0$. Besides four different loss functions, disparity maps are regressed at four scales. At each scale all loss functions are evaluated for both left and right disparity maps. The full objective combines losses from all four scales \textit{s}:
\begin{equation}
    	\mathcal{L}_{total} = \sum_{s=0}^{3} \sum_{d \in \{d_{left}, d_{right} \}} \mathcal{L}_{(s, d)} \\
\end{equation}
\begin{equation}
      \mathcal{L}_{(s, d)} = \alpha_{L1}\mathcal{L}_{L1}+\alpha_{S}\mathcal{L}_{S}+\alpha_{LR}\mathcal{L}_{LR} + \alpha_{DISP}\frac{1}{2^{s}}\mathcal{L}_{DISP},
\end{equation}
where each of the $\alpha$s are the loss weights that should be automatically weighted using one of the loss weighting schemes. A total of 32 losses is used to train the networks. See supplementary material for full details.

\paragraph{Dataset \& Implementation Details}
Depth estimation is evaluated on the KITTI dataset \cite{Geiger2012CVPR, Geiger2013IJRR} using the Eigen split \cite{eigen2014depth} and on the main split of the CityScapes dataset \cite{Cordts2016Cityscapes}.
For all methods, images are down-sampled to a resolution of 256x512 and fed to an encoder-decoder network using batch normalisation \cite{ioffe2015batch}. The encoder is based on a ResNet50 \cite{he2016deep} network; the decoder alternates bilinear interpolation up-sampling and convolutional layers \cite{godard2017unsupervised}. 
The models are trained for 100 epochs on CityScapes and for 30 epochs on KITTI, using an Adam optimiser with a learning rate of \textit{1e-4} (selected based on related works).
For quantitative evaluation, a set of common metrics is used \cite{eigen2014depth, godard2017unsupervised}: Absolute Relative Distance (ARD), Squared Relative Distance (SRD), Root Mean Squared Error (RMSE), log Root Mean Squared Error (log RMSE), and multiple accuracies $\delta$ within a threshold \textit{t} ($\delta_{t}$, with \textit{t} $\in \{1.25,1.25^{2},1.25^{3}\}$). 
It is common for models that are trained on CityScapes to be evaluated on KITTI, since the quality of the ground truth disparities for CityScapes is relatively low. In this paper, this commonly-used evaluation strategy is also used. It means models that are trained on KITTI or on CityScapes are all evaluated on the improved Eigen test split \cite{eigen2014depth} of the KITTI dataset.

\begin{table*}[t]
    \centering
    \ra{1.05}
    \resizebox{1.0\textwidth}{!}{
        \begin{tabular}{@{\extracolsep{4pt}}llccccccc@{\extracolsep{4pt}}}
        \toprule
        && ARD & SRD & RMSE  & RMSE log & $\delta < 1.25$ & $\delta < 1.25^{2}$ & $\delta < 1.25^{3}$ \\ 
        \cline{3-6} \cline{7-9}
	    && \multicolumn{4}{c}{\small{lower is better}} & \multicolumn{3}{c}{\small{higher is better}} \\ 
        \midrule
        \multirow{4}{*}{\rotatebox[origin=c]{90}{\parbox[c]{1.7cm}{\centering Single-Loss}}} 
        & DISP            & 2.0301 & 103.0575 & 28.9682 & 2.4904 & 0.0024 & 0.0055 & 0.0102 \\
        & LR              & 0.7424 & 11.1500 & 17.3990 & 1.8043 & 0.0110 & 0.0382 & 0.0978 \\
        & L1              & 0.2502 & 3.2033 & 7.3217 & \sr{0.3290} & 0.6838 & 0.8650 & \sr{0.9382} \\
        & SSIM            & \tr{0.2002} & \sr{1.6460} & \sr{6.4964} & 0.3682 & \sr{0.7066} & \tr{0.8678} & 0.9308 \\
        \midrule
        \multirow{3}{*}{\rotatebox[origin=c]{90}{\parbox[c]{1cm}{\centering Multi-Loss}}}
        & Equal           & 0.2106 & 1.7460 & \tr{6.6303} & 0.3838 & 0.6887 & 0.8580 & 0.9231 \\
        & Hand-tuned      & \fr{0.1955} & \fr{1.6028} & \fr{6.3803} & \tr{0.3507} & \fr{0.7182} & \fr{0.8734} & \tr{0.9336} \\
        & \ourmethodbf       & \sr{0.1988} & \tr{1.6673} & 6.6989 & \fr{0.3118} & \tr{0.6988} & \sr{0.8723} & \fr{0.9426} \\
        \bottomrule
        \end{tabular}
    }
    \caption{Performance of depth estimation models trained on CityScapes. As is common for CityScapes models, evaluation is performed on the improved Eigen test set \cite{eigen2014depth} in \textbf{depth} (meters). The \fr{first}, \sr{second}, and \tr{third} best scoring methods are highlighted for each metric. 
    }
    \label{tab:depth-cityscapes-results}
\end{table*}

\paragraph{Loss Statistic Estimation} As was mentioned above, a potential downside of using a full history of losses to compute the statistics might result in over-smoothing the estimates. See for example the update steps in~\autoref{eq:update-mean}. To verify whether this is actually the case, different parameters for \textit{t} are tested. Results are depicted in~\autoref{tab:preliminary-cityscapes-statistics}. It is concluded that there is not a significant and apparent effect of using any particular factor \textit{t}. Other initial experimentation confirms this conclusion. Consequently, a full history of loss statistics is kept throughout training.

\paragraph{CityScapes}
In this set of experiments, \ourmethod is compared against the models trained using a single-loss (\textit{e.g.} SSIM or L1) and against the models trained on equal \& hand-tuned weighted combination of losses. For training the models the CityScapes dataset is used, while evaluation is performed on the improved Eigen test split \cite{eigen2014depth}. 
The results are shown in~\autoref{tab:depth-cityscapes-results}.

The performances of the single-loss models show that using auxiliary losses only will not work well for the original task.
For example, the LR loss is tasked with rewarding symmetry in left and right disparity predictions; training with only the LR loss results in predicting purely zero-valued disparity maps, a perfect symmetry, albeit not valuable for depth estimation. These auxiliary losses by themselves do not correctly estimate depth, however they could guide or regularise the learning process.

The performance of using only the SSIM loss is close to the hand-tuned multi-loss counterpart. This is in line with the conclusions reported in ~\cite{groenendijk2020benefit}. 
However, the hand-tuned variant always outperforms the single SSIM loss model and the equal weighted variant.

The performance of \ourmethod is close to the hand-tuned weights for most of the metrics, and even outperforms these for RMSE log and $\delta < 1.25^3$. 
Moreover \ourmethod outperforms equal weighting on all but one metric (RMSE). Combined this shows that \ourmethod can adaptively set good weights for the different loss components. 

\begin{table*}[t]
    \centering
    \ra{1.05}
    \resizebox{1.0\textwidth}{!}{
        \begin{tabular}{@{\extracolsep{4pt}}llccccccc@{\extracolsep{4pt}}}
        \toprule
        && ARD & SRD & RMSE  & RMSE log & $\delta<1.25$ & $\delta<1.25^{2}$ & $\delta<1.25^{3}$ \\ 
        \cline{3-6} \cline{7-9}
	    && \multicolumn{4}{c}{\small{lower is better}} & \multicolumn{3}{c}{\small{higher is better}} \\ 
        \hline
        \multirow{4}{*}{\rotatebox[origin=c]{90}{\parbox[c]{1.7cm}{\centering Single-Loss}}}
        & DISP            & 0.9040 & 13.5185 & 18.3165 & 2.5593 & 0.0000 & 0.0000 & 0.0000 \\
        & LR              & 6.3400 & 442.7141 & 65.5198 & 1.9283 & 0.0097 & 0.0265 & 0.0507 \\
        & L1              & 0.1132 & 0.8551 & 4.4699 & 0.1708 & 0.8700 & 0.9664 & 0.9897 \\
        & SSIM            & \fr{0.0901} & \sr{0.5202} & \fr{3.8581} & \sr{0.1435} & \sr{0.8991} & \tr{0.9755} & \tr{0.9935} \\
        \midrule
        \multirow{9}{*}{\rotatebox[origin=c]{90}{\parbox[c]{2cm}{\centering Multi-Loss}}}
        & \multicolumn{8}{a}{Static Weights} \\
        & Equal           & \tr{0.0923} & \tr{0.5477} & 3.9856 & \tr{0.1464} & \tr{0.8971} & \sr{0.9763} & \sr{0.9936} \\
        & Hand-tuned      & 0.0944 & 0.5609 & \sr{3.8678} & 0.1469 & 0.8950 & 0.9731 & 0.9926 \\
        & \multicolumn{8}{a}{Dynamic Weights} \\
        & Uncertainty \cite{kendall2018multi}       & 1.0712 & 18.5936 & 11.7052 & 0.7417 & 0.3226 & 0.4760 & 0.5989 \\
        & GradNorm \cite{chen2018gradnorm}     & 0.1152 & 1.1530 & 4.7621 & 0.1759 & 0.8669 & 0.9642 & 0.9887 \\
        & Multi-objective \cite{sener2018multi}     & 6.3400 & 442.7141 & 65.5198 & 1.9283 & 0.0097 & 0.0265 & 0.0507 \\
        & \ourmethodbf     & \sr{0.0912} & \fr{0.5108} & \tr{3.8717} & \fr{0.1425} & \fr{0.9023} & \fr{0.9775} & \fr{0.9942} \\
        \bottomrule
        \end{tabular}
    }
    \caption{Performance of depth estimation models trained on KITTI. Evaluated on the improved Eigen test set \cite{eigen2014depth} in \textbf{depth} (meters). The \fr{first}, \sr{second}, and \tr{third} best scoring methods are highlighted. \ourmethod outperforms all other methods on 5 out of 7 metrics.}
    \label{tab:depth-kitti-results}
\end{table*}

\begin{table*}[t]
    \centering
    \ra{1.05}
    \resizebox{1.0\textwidth}{!}{
        \begin{tabular}{@{\extracolsep{4pt}}lcccccc@{\extracolsep{4pt}}}
        \toprule
        & Equal & Hand-tuned & Uncertainty \cite{kendall2018multi} & GradNorm \cite{chen2018gradnorm} & Multi-objective \cite{sener2018multi} & \ourmethodbf \\
        \cline{2-3} \cline{4-6} \cline{7-7}
        \textbf{Equal}              & - & 0.4663 & 1.0000 & 0.8359 & 1.0000 & 0.4463 \\
        \textbf{Hand-tuned}         & 0.5322 & - & 1.0000 & 0.8574 & 1.0000 & 0.4356 \\
        \midrule
        \bf{\ourmethod}             & 0.5537 & 0.5583 & 1.0000 & 0.8681 & 1.0000 & - \\
        \bottomrule
        \end{tabular}
    }
    \caption{Win rates for methods compared to either an equal weighting or a hand-tuned baseline. Evaluated on the Eigen \cite{eigen2014depth} split of KITTI. Models in rows are tested against the models in the columns. For example, \ourmethodbf outperforms equal weighting in 55\% of the cases.}
    \label{tab:depth-kitti-winrates}
\end{table*}

\paragraph{KITTI}
In this set of experiments, \ourmethod is also compared against dynamic baselines developped for multi-task learning: Uncertainty weighing \cite{kendall2018multi}, GradNorm \cite{chen2018gradnorm}, and Multi-objective optimisation \cite{sener2018multi}.
For this experiment the models are trained on the KITTI data using the improved Eigen train and test split. 
The depth estimation results are shown in~\autoref{tab:depth-kitti-results}.

We observe that the single-loss model trained on SSIM is a strong baseline, this is even more distinct than in the CityScapes experiment in~\autoref{tab:depth-cityscapes-results}. 
Moreover, the hand-tuned weights do not outperform SSIM and equal weighting on all metrics. Most likely the set of hand tuned weights from~\cite{godard2017unsupervised} are tuned for a (slightly) different setting and do not generalise to this setting. This shows the importance of finding (optimal) weights for the task and dataset at hand.

From the dynamic weight methods, two of the baselines (Uncertainty weighting and Multi-Objective optimisation) have difficulties training for this task and have far from good evaluation results.
The third baseline, GradNorm, is able to train a decent model, but the performance stays behind compared to the model trained on SSIM, L1 or the model using equal weighted combination.
Finally, \ourmethod shows best performance on 5 from the 7 metrics.

In~\autoref{fig:weights-cs-training}, the loss weights throughout training are shown for hand-tuned weights, GradNorm, Multi-objective optimisation, and CoV-Weighting for training on the CityScapes dataset. It is clear that Multi-objective optimisation underestimates the importance of SSIM and/or L1.
GradNorm assigns the most loss weight to the disparity gradient loss, probably because it is the loss with the lowest magnitude. After the disparity loss, GradNorm assigns most weight to the L1 loss, \textit{c.f.}~\autoref{fig:weights-cs-training}, this is reflected by the performance, in ~\autoref{tab:depth-kitti-results}, the performance of GradNorm is just below the performance of using the L1 loss alone.
\ourmethod attach relatively low importance to SSIM and high importance to the L1 loss compared to the static weights. The weights assigned to SSIM and L1 develop similarly. This is not surprising, since both losses measure reconstructed image quality, whereas the other two losses measure disparity quality. Also it gradually assigns more weight to SSIM and L1, and less to LR.

\paragraph{Win rates}
In this experiment we evaluate the win rate~\cite{zamir2018taskonomy}, the win rate indicates for how many images in the test-set a method is beneficial compared to a baseline method.
This is implemented as a majority voting scheme over the 7 metrics evaluated on a single image.
We show the win rates in~\autoref{tab:depth-kitti-winrates} of the dynamic methods compared to the static multi-loss baselines. 
Since the win rate is the percentage of images in the test set for which a method outperforms another (baseline) method, it gives another insight in the usefulness of the proposed method.
From the results we observe that the three multi-task baselines are outperformed by the equal weighted and hand-tuned weighted variants. 
\ourmethod is favoured by over 55\% of the images in the test set over both equal weighted and hand-tuned weighted models. This shows that our proposed method is able to improve over the performance of hand-tuned models without manually having to tune the loss weights.

\subsection{Semantic Segmentation}
The method is further evaluated on a semantic segmentation task to verify how well it generalises to other tasks. To this end, all loss weighing methods are implemented in conjunction with a Context Encoding Network (EncNet) \cite{zhang2018context}. The authors of \cite{zhang2018context} propose a Context Encoding module, that jointly learns to predict the presence of semantic classes in an image as well as the actual pixel-level class predictions. This module leverages global contextual information to aid pixel-level prediction. Additionally, it is possible to add another head to the penultimate layer of the encoder network to further aid prediction. In total the objective function consists of a standard Cross-Entropy loss (CE), a Semantic Encoding loss (SE), and an auxiliary Cross-Entropy loss (AUX) from a separate Fully Convolutional (FCN) head. The hand-tuned parameters $\alpha$ as given in \cite{zhang2018context} are $\{\alpha_{CE}=1.0, \alpha_{SE}=0.2, \alpha_{AUX}=0.2\}$ Again, these weights are normalised to ensure $\sum_{i} \alpha_{i}=1.0$

\begin{wraptable}{l}{.44\textwidth}
    \centering
    \begin{tabular}{lcc}
        \toprule
        & \bf pACC & \bf mIoU \\
        \hline
        \multicolumn{3}{a}{Single-Loss} \\
        CE                                          & {0.769} & {0.440} \\
        AUX                                         & 0.012 & 0.004 \\
        SE                                          & 0.010 & 0.001 \\
        \hline
        \multicolumn{3}{|a}{Static Weights} \\
        Equal                                   & 0.759 & 0.419  \\
        Hand-tuned                              & {0.768} & {0.437}\\
        \multicolumn{3}{|a}{Dynamic Weights} \\
        Uncertainty \cite{kendall2018multi}       & \fr{0.781} & \sr{0.448} \\
        GradNorm \cite{chen2018gradnorm}          & 0.750 & 0.404 \\
        Multi-objective \cite{sener2018multi}     & 0.012 & 0.003 \\
        \hline
        \multicolumn{3}{|a}{Proposed} \\
        \ourmethodbf ($\gamma=10^{-2}$)            & \sr{0.779} & \fr{0.455}  \\
        \ourmethodbf ($\gamma=10^{-3}$)            & \tr{0.770} & \tr{0.441}  \\
        \ourmethodbf ($\gamma=10^{-4}$)            & {0.768} & 0.436  \\
        \bottomrule
    \end{tabular}
    \caption{Performance of semantic segmentation models trained on PASCAL Context. Evaluated on 5104 images in the validation set.
    The \fr{first}, \sr{second}, and \tr{third} best scoring methods are highlighted for each evaluation metric.
    \ourmethodbf is top performing when suitable learning-rate is chosen.
    }
    \label{tab:semantics-pcontext-results}
\end{wraptable}

\paragraph{Implementation Details}
The EncNet is adapted and augmented with all loss weighing methods. The methods are tested on the PASCAL Context dataset \cite{mottaghi2014role, pascal-voc-2010} which uses 4998 training images and 5105 validation images. For each image, there are annotations for up to 59 semantic classes. The encoder network is a ResNet50 \cite{he2016deep} network using batch normalisation \cite{ioffe2015batch}; for the decoders, there is one Encoding Context Module that is attached to the final layer of the encoder, and one FCN head attached to the penultimate layer of the decoder. For optimisation, an SGD optimiser is used with a learning rate of \textit{1e-4}, unless otherwise stated. The network is pre-trained on ImageNet \cite{imagenet_cvpr09} and then trained for 40 epochs on PASCAL Context. For quantitative evaluation, Pixel Accuracy (pACC) and Mean Intersection over Union (mIoU) are used, as in \cite{zhang2018context}. Background pixels are ignored during evaluation.

\paragraph{Results}
In this experiment different models are compared trained on a single loss, on a static weighted combination of losses, and on the dynamic weighted combination.
The results on PASCAL Context are depicted in~\autoref{tab:semantics-pcontext-results}. 
Similarly as in the depth prediction task, also for semantic segmentation there is a single loss which provides a strong baseline.
For this experiment, using only the CE loss yields performance on par with or slightly better than the hand-tuned weights. 
This could be due to the more shallow encoder network used (compared to ~\cite{zhang2018context}) or because the FCN head could be replaced with another Context Encoding module to improve performance as suggested in \cite{zhang2018context}.
For this task, multi-objective weighting yields far from satisfactory results, probably due to attempting to assign high weight to an auxiliary loss with high gradient magnitude that has negative transfer with the main loss. GradNorm on the other hand, performs slightly worse than equal weighting. Similar to the case of multi-objective training, a possible explanation could be that a loss that serves purely as an auxiliary loss receives a too high loss weighting because its gradient has a high magnitude.
\begin{table*}[t]
    \centering
    \ra{1.05}
    \resizebox{1.0\textwidth}{!}{
        \Huge
        \begin{tabular}{ccc@{\hspace{0cm}}c@{\hspace{0cm}}c}
        \toprule
        \textbf{Variant} & RMSE $\downarrow$ & $\delta<1.25 \uparrow$ & \textbf{Loss Weights} & \textbf{Scale Weights} \\
        \midrule \addlinespace[3pt]
        $\frac{\sigma_L}{\mu_L}$        & \fr{6.6483} & 0.7050 & \noindent\parbox[c]{\hsize}{\includegraphics[width=1.05\textwidth]{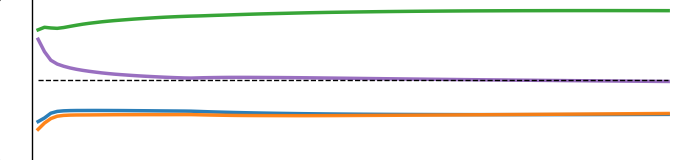}} & \noindent\parbox[c]{\hsize}{\includegraphics[width=1.05\textwidth]{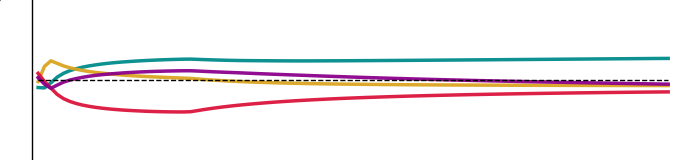}} \\ \addlinespace[3pt]
        $\frac{\mu_L}{\sigma_L}$        & 6.6977 & 0.7043 & \noindent\parbox[c]{\hsize}{\includegraphics[width=1.05\textwidth]{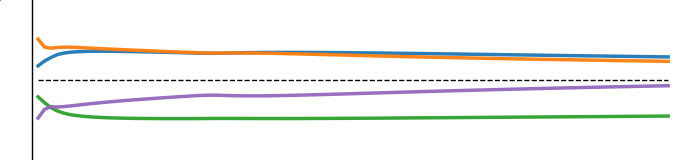}} & \noindent\parbox[c]{\hsize}{\includegraphics[width=1.05\textwidth]{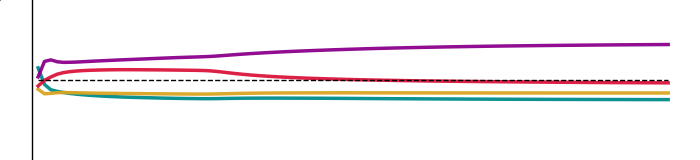}} \\ \addlinespace[3pt]
        $\frac{\sigma_l}{\mu_l}$ $^*$     & \sr{6.6556} & \sr{0.7063} & \noindent\parbox[c]{\hsize}{\includegraphics[width=1.05\textwidth]{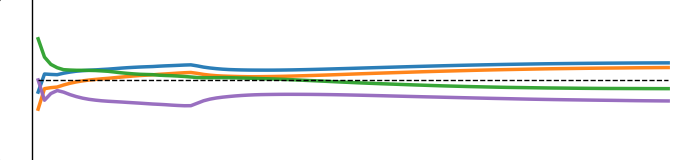}} & \noindent\parbox[c]{\hsize}{\includegraphics[width=1.05\textwidth]{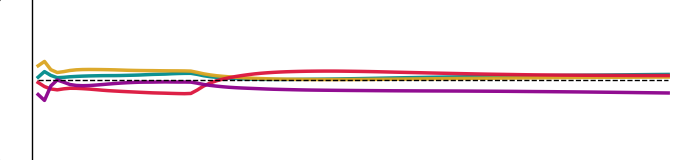}} \\ \addlinespace[3pt]
        $\frac{\mu_l}{\sigma_l}$        & 6.7670 & \fr{0.7072} & \noindent\parbox[c]{\hsize}{\includegraphics[width=1.05\textwidth]{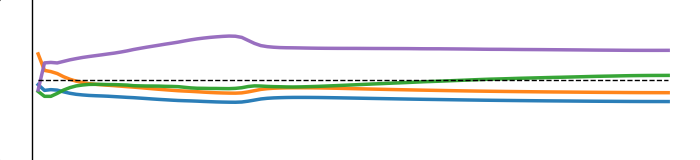}} & \noindent\parbox[c]{\hsize}{\includegraphics[width=1.05\textwidth]{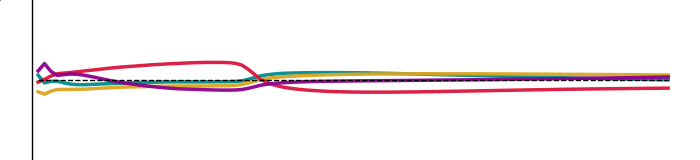}} \\ \addlinespace[3pt]
        &&& \noindent\parbox[c]{0.8\textwidth}{\includegraphics[width=0.9\textwidth]{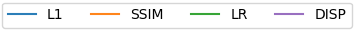}} & \noindent\parbox[c]{0.8\textwidth}{\includegraphics[width=0.9\textwidth]{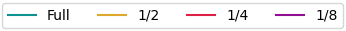}} \\
        \bottomrule
        \end{tabular}
    }
    \caption{A comparison of four variants of \ourmethod on the CityScapes dataset. The top two rows indicate methods that have been trained using statistics over the \textbf{losses}; the bottom two rows indicate methods using statistics over the  \textbf{loss ratios}. The $^*$ indicates \ourmethodnospace as used throughout the paper. The \fr{first} and \sr{second} best scoring methods are highlighted for each metric. For each method the mean loss weights over the four different losses and mean loss weights over the different scales are shown; the weights are shown in the range [0, 0.5]; the dotted horizontal lines indicate a weight of 0.25. Best seen in color.}
    \label{tab:depth-cityscapes-weights-comparison}
    \vspace{-2mm}
\end{table*}

Uncertainty weighting and \ourmethod out perform both equal weighting and hand-tuned weights. For uncertainty weighting, it seems that this is in part due to the unnormalised weights used, \textit{i.e.} $\sum_{i} \alpha_{i}$ can have any value, which in turn tunes the global learning rate to a higher value. Similarly, \ourmethod also benefits from a higher learning rate for this task ($\gamma=10^{-2}$). {Choosing a suitable learning rate is necessary regardless of the method that is chosen; our method does not require additional learning rate tuning compared to any of the baseline methods.}

\subsection{\ourmethod Variations}
In this final set of experiments, several variants of \ourmethodnospace are explored. So far, the loss weights have been derived from the mean and variance in the observed loss ratios $\tfrac{\sigma_l}{\mu_l}$. In this experiment the mean and variances from the observed losses  are used: $\tfrac{\sigma_L}{\mu_L}$, and the inverse-weighting is used: $\tfrac{\mu_l}{\sigma_l}$ \& $\tfrac{\mu_L}{\sigma_L}$.
The rationale for the inverse-weighting is to test the hypothesis that low-variance losses should be assigned a low weight.

For this experiment the depth estimation task on the CityScapes dataset is used. The performance is measured in RMSE and $\delta<1.25$. The results are in~\autoref{tab:depth-cityscapes-weights-comparison}. 
For qualitative analysis also the weights over time for the different losses (\emph{middle}) and the different scales (\emph{right}) are plotted. 

From the results we first observe that all variants obtain decent performance, albeit assigning different weights to each of the losses and each of the scales. From this we may conclude that for the multi-loss single task scenario multiple paths lead to good performance.

Based on the previous experiments the expectation is that models with relatively high weights to L1 and/or SSIM will perform the best. \ourmethod and the $\tfrac{\mu_L}{\sigma_L}$-variant do so. However, per scale these methods assign different weights, since final predictions are required at full scale, the coarser scales act as auxiliary losses. The variant $\tfrac{\mu_L}{\sigma_L}$ wrongly assigns high weights to losses at coarse scale, \textit{c.f.}~\autoref{tab:depth-cityscapes-weights-comparison}.

\section{Conclusion} \label{sec:conclusion}
In this paper, \ourmethod has been introduced to automate tuning of loss weights specifically on single-task problems. 
Related methods from multi-task learning \cite{sener2018multi, kendall2018multi, chen2018gradnorm}, are shown to not always be suited in a single-task setting, given that auxiliary losses cannot be weighed too heavily because they by themselves do not solve the task. These losses are often comparatively small and less complex. Consequently, the losses show less variance throughout training. \ourmethod explicitly makes use of these statistics, inspired by the coefficient of variation and assigns higher weights to losses that show higher relative variance.
Experimentally \ourmethod either outperforms or performs on par with hand-tuned defined weights and outperform these when the optimal learning rate is used.

Future work could address \ourmethod in a multi-task setting. 
The use of the coefficient of variation allows to compare different one another. Since this holds also for losses from different tasks, it would be interesting to see how \ourmethod performs in multi-task learning.

\bibliographystyle{unsrt}
\bibliography{references}

\end{document}